\documentclass[runningheads]{llncs}
\usepackage{graphicx}

\usepackage[american]{babel}
\usepackage{amsmath}
\usepackage{booktabs}   
\usepackage{multirow}
\usepackage{dblfloatfix}
\usepackage{multicol}
\usepackage{makecell}%
\usepackage{natbib} 
    \setcitestyle{square}
    \setcitestyle{numbers}
    \setcitestyle{comma}
    
\usepackage{mathtools} 
\usepackage{booktabs} 
\usepackage{tikz} 

\begin{document}
\title{Neural Additive Vector Autoregression Models for Causal Discovery in Time Series}
%
%
\author{Bart Bussmann\inst{1} \and
Jannes Nys\inst{1} \and
Steven Latré\inst{1}}
\authorrunning{B. Bussmann et al.}
\titlerunning{Neural Additive Vector Autoregression for Causal Discovery in Time Series}

\institute{IDLab, University of Antwerp - imec, Antwerpen, Belgium \\ \email{\{firstname.lastname\}@uantwerpen.be}}

\maketitle              
\begin{abstract}
Causal structure discovery in complex dynamical systems is an important challenge for many scientific domains. Although data from (interventional) experiments is usually limited, large amounts of observational time series data sets are usually available. Current methods that learn causal structure from time series often assume linear relationships. Hence, they may fail in realistic settings that contain nonlinear relations between the variables. We propose Neural Additive Vector Autoregression (NAVAR) models, a neural approach to causal structure learning that can discover nonlinear relationships. We train deep neural networks that extract the (additive) Granger causal influences from the time evolution in multi-variate time series. The method achieves state-of-the-art results on various benchmark data sets for causal discovery, while providing clear interpretations of the mapped causal relations. 
\end{abstract}

\section{Introduction}
Discovering mechanisms and causal structures is an important challenge for many scientific domains. Randomized control trials may not always be feasible, practical or ethical, such as in the domain of climate sciences and genetics. Therefore, when no interventional data is available, we are forced to rely on observational data only. 

In dynamical systems, the arrow of time simplifies the analysis of possible causal interactions in the sense that we can assume that only preceding signals are a potential cause of the current observations. A common approach is to test time-lagged causal associations in the framework of Granger causality \citep{granger}. These methods often model the time-dependence via linear causal relationships, with Vector AutoRegression (VAR) models as the most common approach. 

Even though there is extensive literature on nonlinear causal discovery (e.g. \citep{marinazzo2011nonlinear, stephan2008nonlinear}) relatively few others (e.g. \citep{neuralgranger, statisticalRNN}) have harnessed the power of deep learning for causal discovery in time series. These methods operate within the Granger causality framework and use deep neural networks to model the time dependencies and interactions between the variables. In principle, deep learning approaches make it possible to model causal relationships, even when they are nonlinear.  
While these methods have a high degree of expressiveness, this flexibility comes at a cost: interpretation of the causal relations learned by black-box methods is hindered, while this is essentially the goal of causal structure learning. To overcome this, these methods learn to set certain input weights to zero, which they interpret as an absence of Granger Causality.

In this work, we propose the Neural Additive Vector Autoregression (NAVAR) model to resolve this problem. NAVAR assumes an additive structure, where the predictions depend linearly on independent nonlinear functions of the individual input variables. We model these nonlinear functions using neural networks. In comparison to other works using Granger causality for causal discovery in time series, our work differs in the following ways:
\begin{enumerate}
    \item Compared to common linear methods, our method can easily capture (highly) nonlinear relations.
    
    \item While being able to model nonlinear relations, NAVAR maintains a clear interpretation of the causal dependencies between pairs of variables. In contrast to other deep learning methods that  resort to feature importance methods, NAVAR uses the interpretational power of additive models to discover Granger causal relationships.

    \item By using an additive model of learned transformations of the input variables, our model allows not only for the discovery of causal relationships between pairs of time series but also inspection of the functional form of these causal dependencies. Thanks to the additive structure, we can inspect the direct contribution of every input variable to every output variable.

    \item The additive structure allows us to score and rank causal relations. Since we can compute the direct contribution of each input variable to each output variable independently, the variability of these contributions can be used as evidence for the existence of a causal link.
\end{enumerate}

The rest of this paper is structured as follows: Section 2 introduces the Granger causality framework and VAR models. In Section 3 we generalize this notion to the additive nonlinear case and introduce NAVAR models that can estimate Granger causality using neural networks. In Section 4 we evaluate the performance of NAVAR on various benchmarks and compare it to existing methods. Finally, in Section 5 we discuss related work and in Section 6 we conclude and discuss directions for future work.

\section{Granger Causality and the VAR Model}
Let $X_{1:T}  = \{ X_{1:T}^{(1)}, X_{1:T}^{(2)}, .., X_{1:T}^{(N)} \}$ be a multivariate time series with $N$ variables and $T$ time steps.  Our goal is to discover the causal relations between this set of time series.
(Pairwise) Granger causality is one of the classical frameworks to discover causal relationships between time series. In this framework, we model the time series as:
\begin{equation}
    X^{(i)}_t = g^i(X^{(1)}_{<t}, ..., X^{(N)}_{<t}) + \eta^i_t
\end{equation}
where $X^{(i)}_{<t} = X^{(i)}_{1:t-1}$ denotes the past of $X^{(i)}$, and and $\eta_t$ is an independent noise vector.
A variable $X^{(i)}$ is said to Granger cause another variable $X^{(j)}$ if the past of the set of all (input) variables $\{X^{(1)}_{<t},...,X^{(i)}_{<t},  ..., X^{(N)}_{<t}\}$ allows for better predictions for $X^{(j)}_t$ compared to the same set where the past of $X^{(i)}$ is not included: $\{X^{(1)}_{<t},...,X^{(i-1)}_{<t},X^{(i+1)}_{<t},  ..., X^{(N)}_{<t}\}$. Granger causality approaches assume causal sufficiency. We refer to the directed graph with the variables $X^{(i)}$ as vertices, and links representing Granger causality between two variables as the Granger causal graph.

 In the VAR framework, the time series $X^{(j)}_t$ is assumed to be a linear combination of all past values (up to some maximum lag $K$) and independent noise term. This means that every value $X^{(j)}_t$ can be modeled as: 

\begin{equation}
    X_{t}^{(j)}  = \beta^j + \sum\limits^{N}_{i=1}\sum\limits^{K}_{k=1} [A_k]^{ij} X^{(i)}_{t-k}   + \eta_t^{j}
\end{equation}

Where $A_k$ is a $N \times N$ time-invariant matrix which identifies the interaction between the variables, $\beta$ is a $N$-dimensional bias vector, and $\eta_t$ is an independent noise vector with zero mean. A common approach to infer which pairs of variables are \textit{not} Granger causal is to identify $i$ and $j$ for which $[A_k]^{ij} = 0$ for all time lags $k=1, ..., K$.

\section{NAVAR: Neural Additive Vector AutoRegression}

The idea underlying the linear VAR model is simple and it can be surprisingly effective. For instance, in the NeurIPS 2019 Causality for Climate competition, the winners used four variations based on the standard linear VAR model \citep{Cocola}. However, a limitation of the VAR model is that it can only model linear interactions. Guided by the success and reliability of VAR models for Granger-causal discovery, in this work, we generalize the VAR model to allow for nonlinear additive relationships between variables:
\begin{equation}\label{eq:nonlinVAR}
    X_t^{(j)} =  \beta^j + \sum\limits_{i=1}^N f^{ij}(X^{(i)}_{t-K:t-1}) + \eta_t^{j}
\end{equation}
Here, $f^{ij}$ is a nonlinear function describing the relationship between the past $K$ values of $X^{(i)}$ on the current value of $X^{(j)}$. 
Note that the VAR model is the special case where $f^{ij}$ is linear. We can identify Granger causality in the following way: if variable $X^{(i)}$ is not a Granger cause of another variable $X^{(j)}$ then $f^{ij}$ is invariant to the values of $X^{(i)}_{t-K:t-1}$. In other words, if $f^{ij}$ is a constant function of \emph{all} values $X^{(i)}_{t-K:t-1}$, then $X^{(i)}$ is not a Granger cause of $X^{(j)}$. 

The choice of this additive model is built on the following assumption. In many practical applications, the functional dependence of a variable $X^{(i)}_t$ on the history of a variable $X^{(i)}_{<t}$ is complex, with e.g. nonlinear functions across multiple time lags. However, dependencies on multiple time series can usually be well approximated by additive models. Therefore, we introduce an additive structure for the contributions stemming from the different variables, but do not impose an additive restriction to contributions from different time lags.

\begin{figure}[!b]
    \centering
    \includegraphics[width=0.63\linewidth]{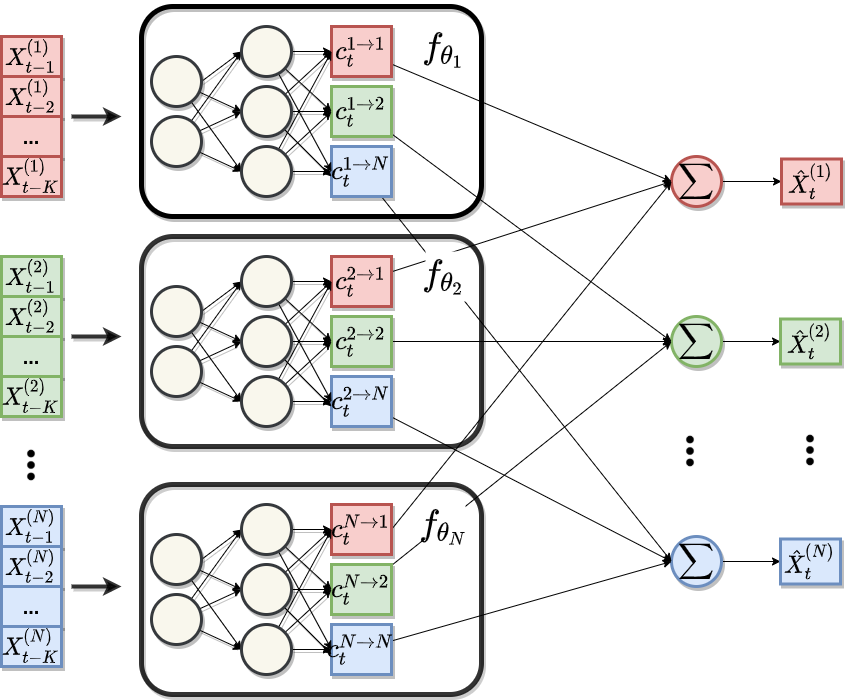}
    \caption{Graphical representation of the NAVAR model with MLPs. For every time step $t$ and every variable $X^{(i)}$, we compute a nonlinear combination of its past $X^{(i)}_{t-K:t-1}$ (with a maximum time lag $K$) as the contribution to every other variable. To compute the estimate of $X^{(j)}_t$, all contributions $c_{t}^{i \to j}$ are summed.}
    \label{fig:causalnet}
\end{figure}

We choose to use deep neural networks (DNN) to model the nonlinear function $f^{ij}$. In our method, dubbed Neural Additive Vector AutoRegression (NAVAR), we train separate models on the past of each variable to predict its contribution to the value of all variables at the next time step. In particular, at every time step $t$, we pass the past values of a variable $X^{(i)}$ to a neural network $f$ with $N$ output nodes to compute its contribution to all other variables $X^{(j)}$:
\begin{equation}\label{eq:contributions}
    c_{t}^{i  \rightarrow j } = [f_{\theta_{i}}(X^{(i)}_{t-K:t-1} )]^j
\end{equation}
The function $[f_{\theta_i}]^j$ is the $j$th output of the neural network $f$ with parameters $\theta_{i}$. A graphical overview of the method can be found in Figure \ref{fig:causalnet}.  In principle, one can choose a wide variety of neural networks for $f$, e.g. Multi-Layered Perceptrons (MLP), Recurrent Neural Networks (RNN), and Convolutional Neural Networks (CNN). In our experiments, we consider MLPs and LSTMs  \citep{gers2000learning} to demonstrate the concept, since the additive structure is key to its success. In the LSTM version of our model, single time steps of a variable are sequentially passed to the networks, and thus the networks predict the contributions based only on $X_{t-1}$ and its recurrent hidden states (in contrast to K inputs to the MLP). Therefore, the size of the LSTM network does not increase for larger lags and is thus particularly scalable to longer lags. Although these backbones already outperform the state of the art, we envision that more complex backbone architectures for $f$ could potentially further increase performance.

The resulting prediction for $X^{(j)}$ at time $t$ is the sum of all its incoming contributions: 
\begin{equation}\label{eq:incomingcontr}
    \hat{X}^{(j)}_{t} = \beta^j + \sum\limits_{i=1}^N c_{t}^{i \rightarrow j}
\end{equation}

We choose this additive structure of neural networks as it is a natural extension of the VAR framework with nonlinearities (see Equation \ref{eq:nonlinVAR}) and it allows us to uncover the causal links from $X^{(i)}$ to $X^{(j)}$ by inspecting the direct contributions $c_{t}^{i \rightarrow j }$. Granger causality requires us to estimate the predictions for $X^{(j)}_t$ when the past of $X^{(i)}$ is not included, which in our framework can be directly obtained by ignoring the corresponding contribution $c_{t}^{i  \rightarrow j }$ in equation \eqref{eq:incomingcontr}. This is a key feature of our method that allows it to be scalable: we avoid the necessity to perform multiple fits of a neural network, such as a fit including and excluding the past of variable $X^{(i)}$, when testing the predictive power due to $X^{(i)}$ (see the discussion in Related Work). 

The regression networks are trained using the MSE loss function. We introduce an $l_1$ penalty to the contributions $c_t^{i \to j}$ in order to promote sparsity in the resulting causal link structure. Assuming that large causal networks will have a similar number of causes per variable compared to smaller networks, we choose to penalize the sum of the absolute value of received contributions per variable instead of the mean contribution size. This results in the following loss function for the predictions at a time step $t$:

\begin{equation}
\begin{aligned}
\mathcal{L}_t(\beta, \theta) = &  \frac{1}{N}\sum^{N}_{j=1} \left( \beta^j + \sum^{N}_{i=1}  [f_{\theta_{i}}(X^{(i)}_{t-K:t-1} )]^j  - X^{(j)}_t  \right)^2 \\
      & + \frac{\lambda}{N}\sum\limits_{i,j=1}^N \left| [f_{\theta_{i}}(X^{(i)}_{t-K:t-1} )]^j \right|
\end{aligned}
\end{equation}

Furthermore, we add a weight decay term to the loss with coefficient $\mu$.

 In order to make the contributions comparable, every individual time series is normalized such that it has mean zero and standard deviation one before training. After training the networks, we deduce the causal links from the variability of the contributions in equation \eqref{eq:contributions}.  The rationale to reconstruct the Granger causal graph is that if a certain variable has a large causal influence on another variable, then it will send a large variety of contributions over the course of time. However, if a variable $X^{(i)}$ is not a Granger cause of another variable $X^{(j)}$ then $f^{ij}$ is a constant function, because $X^{(j)}$ is invariant to the values of $X^{(i)}_{t-K:t-1}$. To score a potential causal link $X^{(i)} \to X^{(j)}$ with the trained neural network, we therefore compute the standard deviation of the set of contributions $c_t^{i\to j}$ for all $t \in \{K+1, T\}$:
\begin{equation}\label{eq:score}
    \text{score}(i \to j) = \sigma (\{c_{K+1}^{i \rightarrow j}, c_{K+2}^{i \rightarrow j}, ... ,   c_{T}^{i \rightarrow j}\})
\end{equation}

In all of our experiments, we use the ReLU activation function and the Adam optimizer \citep{kingma2014adam} to train our networks. Our implementation of NAVAR and code to reproduce the experiments can be found at: https://github.com/bartbussmann/NAVAR

\section{Experiments}

\subsection{Interpretable Contributions}

First, we investigate the ability of our model to learn interpretable nonlinear causal dependencies on a toy dataset. We construct the dataset with three variables ($N=3$) and 4000 time steps ($T=4000$) based on the following SCM:

\begin{figure}[!b]
    \centering

    \includegraphics[width=\linewidth]{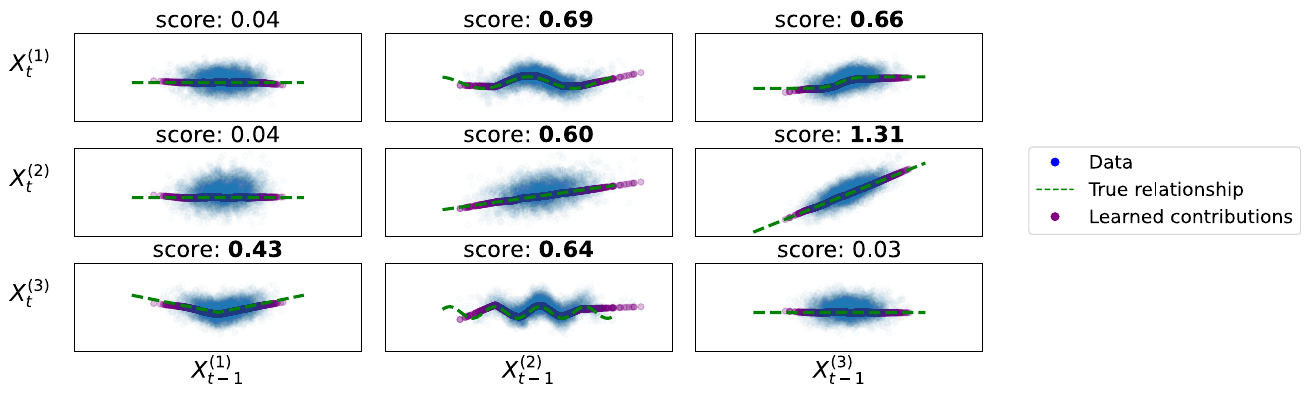}
    \caption{The learned contributions between pairs of variables in our synthetic dataset. The learned contribution functions closely reflect the true causal influence, showing the power of NAVAR models in both Granger causal discovery and interpretability. The causality score from Equation \eqref{eq:score} is given for each potential link. The scores of true causal relationships are presented in boldface. }
    \label{fig:interactions}
\end{figure}

\begin{align}
    X^{(1)}_t &= \text{cos}(X^{(2)}_{t-1}) + \text{tanh}(X^{(3)}_{t-1}) + \eta^{1}_t \notag \\
    X^{(2)}_t &=  0.35 \cdot X^{(2)}_{t-1} + X^{(3)}_{t-1} +  \eta^{2}_t \notag\\
    X^{(3)}_t &= \left|0.5\cdot X^{(1)}_{t-1}\right|  + \text{sin}(2 X^{(2)}_{t-1}) + \eta^{3}_t \notag
\end{align}

where $\eta^i_t \sim \mathcal{N}(0,1)$ for  $i=1,2,3$. 

We train a NAVAR (MLP) model on this dataset and investigate the learned contributions between pairs of variables. In Figure \ref{fig:interactions} we find that the model has learned contributions that are similar to the ground truth causal relationship. Furthermore, we find that for the pairs of variables that are not Granger causal, the learned contribution function has very little variability. This illustrates that our rationale for using the standard deviation of the learned contributions as measure for Granger causal influence is appropriate.

\begin{figure*}[!t]
    \centering
    \includegraphics[width=\linewidth]{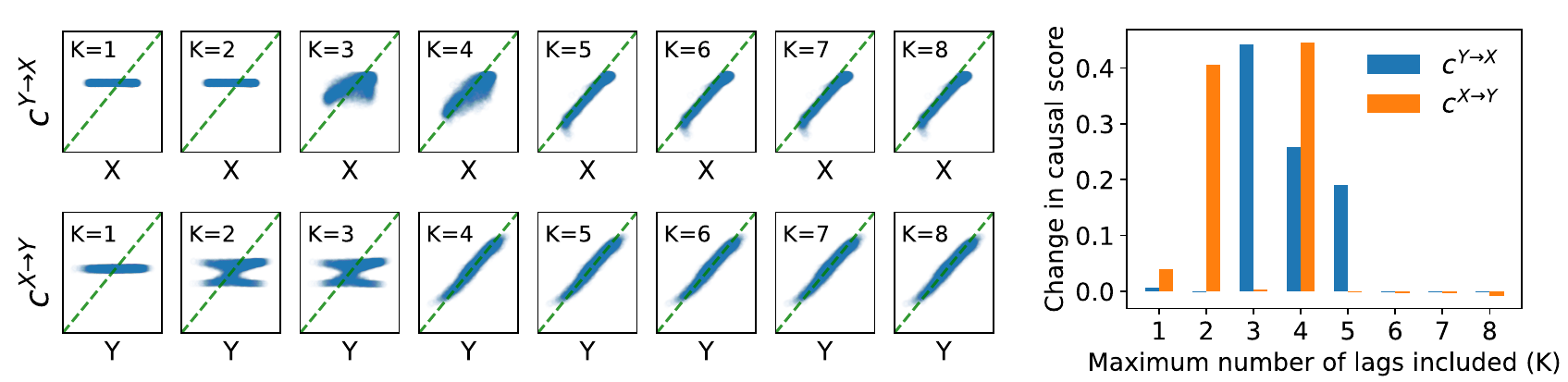}
    \caption{Left Panel: NAVAR discovers coupled nonlinearities within time series across multiple lags. Contributions are shown for different $K$ to study the time lag (by masking the input of the fitted model at higher lags). The diagonal represents learned contributions that perfectly predict the target variable. At lags with true causal relationships, the standard deviation of the contribution increases and the mean squared error (distance to the diagonal) decreases. Right panel: change in causal score from Equation ~\eqref{eq:score} when including $K$ lags, with respect to the case of including $K-1$ lags. For $c^{Y \to X}$, we observe a high causal score contribution at $K=3,4,5$, while for $c^{X \to Y}$ we observe high scores at $K=2,4$, both in agreement with the underlying SCM in Equation~\eqref{eq:synth2}.}
    \label{fig:multipleinteractions}
\end{figure*}

Next, we investigate how to interpret the contributions when the underlying data contains nonlinear interactions across multiple time lags. To this end, we construct a second synthetic dataset with two variables ($N=2$) and $4000$ time steps ($T=4000$) based on the following structural causal model:

\begin{align}\label{eq:synth2}
    X_t &= \text{cos}(Y_{t-3} + Y_{t-4}  + Y_{t-5}) + \eta^{1}_t \notag \\
    Y_t &= X_{t-2} \cdot X_{t-4} +  \eta^{2}_t \notag\\
\end{align}

where $\eta^i_t \sim \mathcal{N}(0,0.1)$ for  $i=1,2$

We train a NAVAR (MLP) model with a maximum lag $K=8$. Although we do not enforce interpretable additive contributions of individual time lags and thus cannot extract the isolated causal influence of individual time lags, we can still investigate the effect of leaving time lags out. Therefore, we mask the input of the fitted model from a certain maximum time lag. In Figure ~\ref{fig:multipleinteractions} three observations can be made: (1) after adding a lag with a true causal link the standard deviation of the contribution increases significantly, which motivated the use of our score function; (2) for time lags with a true causal link the mean squared error decreases; (3) for time lags without a true causal relationship neither of these change significantly, showing that the model did not pick up on spurious contributions (i.e.\ correlations). We point out that the above analysis is made feasible due to the additive structure which allows us to study pairs of variables in isolation from other contributions, and the sparsity penalty that forces the model to consider mostly direct causes.

\subsection{CauseMe - Earth Sciences Data}
We evaluate our algorithm on various datasets on the CauseMe platform \citep{CauseMe}. The CauseMe platform provides benchmark datasets to assess and compare the performance of causal discovery methods. The available benchmarks contain both datasets generated from synthetic models mimicking real challenges, as well as real-world data sets in the earth sciences where the causal structure is known with high confidence. The datasets vary in dimensionality, complexity, and sophistication, and come with various challenges that are common in real datasets, such as autocorrelation, nonlinearities, chaotic dynamics, extreme events, nonstationarity, and measurement errors \citep{runge2019inferring}. On the platform, users have registered over 80 methods for Granger cause discovery. 

We compare our methods with four baseline methods implemented by the platform, namely: VAR \citep{seabold2010statsmodels}, Adaptive LASSO \citep{zou2006adaptive}, PCMCI \citep{runge2018causal}, and FullCI \citep{runge2019detecting}. The VAR and Adaptive Lasso methods are both linear regression methods, where the latter consists of computing several Lasso regressions with iterative feature re-weighting. PCMCI and FullCI are constraint-based methods and perform conditional independence tests. Both of these algorithms come with three different independence tests, namely the linear ParCorr test and the nonlinear GPDC and CMI tests. For these methods, we report the results of the best scoring independence test. Furthermore, we compare NAVAR with SLARAC and SELVAR \citep{Cocola}, the two algorithms that won the NeurIPS 2019 Causality for Climate competition. SLARAC fits a VAR model
on bootstrap samples of the data, each time choosing a random number of lags to include, whereas SELVAR selects edges employing a hill-climbing procedure based on the
leave-one-out residual sum of squares of a VAR model.

\begin{table*}[!b]
\centering
\caption{Average AUROC on various datasets of the CauseMe platform. Performance of the baseline methods Adaptive LASSO, PCMCI, and FullCI are not available for the hybrid and real-world datasets. For each dataset, we provide the total number of time steps $T$ and the number of variables $N$. Datasets with purely linear dynamics are indicated by an asterisk. Models with the highest AUROC are indicated in boldface.}
    \centering
    \small

    \begin{tabular}{lllllllll}
        \toprule
        \multirow{3}{*}{} & \multicolumn{4}{c}{Nonlinear VAR} & Climate* & Weather & River\\
        \cmidrule(lr){2-5}  \cmidrule(lr){6-8}
        {} & N = 3 & N = 5 & N = 10 & N = 20    & N = 40 & N = 10 & N = 12 \\
         & T = 300 & T = 300 & T = 300 & T = 300 &  T = 250 & T = 2000 & T = 4600 \\
        \midrule
         NAVAR (MLP) & 0.86 & \textbf{0.86} & \textbf{0.89} & \textbf{0.89} & 0.80 &  0.89 & \textbf{0.94}\\
          NAVAR (LSTM) & 0.85 & 0.84 & 0.84 & 0.81 & 0.80 &  0.89 & \textbf{0.94}\\

         SELVAR & \textbf{0.88} & \textbf{0.86} & 0.86 &  0.85 & 0.81 & 0.90 & 0.87\\
        
         SLARAC & 0.74 & 0.76 & 0.78 & 0.78 & \textbf{0.95} & \textbf{0.95} & 0.93\\
         VAR & 0.72 & 0.69 & 0.68 & 0.66 & 0.80 &  0.79 &  0.71\\
        
         Ad. LASSO & 0.82 & 0.79 & 0.79 & 0.78 &- & - & -\\
         PCMCI & 0.85 & 0.82 & 0.83 & 0.82 & - & - & -\\
         FullCI & 0.83 & 0.81 & 0.81 & 0.82 & - &- &-\\

        \bottomrule
    \end{tabular}
 \label{table:causeme}
\end{table*}

Every experiment (e.g. Climate, with N=40, T=250) consists of 200 datasets. For every experiment, we tune our hyperparameters (hidden units,  batch size, learning rate, contribution penalty coefficient $\lambda$, and weight decay $\mu$) on the first five datasets, of which we use the first 80\% for training and the final 20\% for validation. The optimal hyperparameters are tabulated in Appendix A\footnote{Appendices and code can be found at https://github.com/bartbussmann/NAVAR}. We set the maximum lag parameter $K$ based on information provided by CauseMe, and train on every dataset for 5000 epochs. The AUROC scores are calculated by the CauseMe platform, where self-links are ignored.

We run our method on the synthetic nonlinear VAR dataset, the hybrid climate and weather dataset, and the real-world river run-off dataset. The results in Table \ref{table:causeme} show that NAVAR (MLP) models outperform the other methods on most of the nonlinear VAR datasets. Interestingly, where the performance of most methods declines as the number of variables $N$ increases, the performance of NAVAR (MLP) does not decrease.
Noting the relative poor performance of SLARAC on the nonlinear VAR dataset compared to its performance on the linear climate dataset, we conclude that this algorithm is very well suited for discovering \textit{exactly linear} relationships. Although NAVAR models might be slightly too flexible for linear datasets, it outperforms the other methods on the real-world river run-off dataset. This strengthens our intuition that many real-world processes can be modeled by an additive combination of nonlinear functions.

\subsection{DREAM3 - Gene Expression Data}

\begin{table*}[!b]
    \centering
    \small
    \caption{Average AUROC on the DREAM3 gene expression dataset. Neural methods are indicated with an asterisk, and their scores are obtained from~\protect\citep{statisticalRNN}. Models with the highest AUROC are indicated in boldface.}
    \begin{tabular}{lllllllll}
        \toprule
         Model & \textbf{E.Coli 1} & \textbf{E.Coli 2} & \textbf{Yeast 1} & \textbf{Yeast 2} & \textbf{Yeast 3} \\ [0.5ex] 
        \midrule
     NAVAR (MLP)* & 0.696 & 0.649 & 0.681  & \textbf{0.601} & 0.594\\ 
         NAVAR (LSTM)* & \textbf{0.715} & \textbf{0.682} & \textbf{0.695}  & 0.599 & \textbf{0.597}\\ 
   cMLP* & 0.644 & 0.568 & 0.585 & 0.506 & 0.528\\ 
    cLSTM* & 0.629 & 0.609 & 0.579 & 0.519 & 0.555\\ 
   TCDF*  & 0.614 & 0.647 & 0.581 & 0.556 & 0.557\\ 
    SRU* & 0.657 & 0.666 & 0.617 & 0.575 & 0.550\\ 
    eSRU* & 0.660 & 0.629 & 0.627 & 0.557 & 0.550\\ 
  SELVAR & 0.551 & 0.536 & 0.556 & 0.516 & 0.534 \\
 SLARAC  & 0.580 & 0.509 & 0.526 & 0.503 & 0.494\\
        \bottomrule
    \end{tabular}
\label{table:dream}
\end{table*}

Next, we evaluate our algorithm on the DREAM3 dataset, a  simulated gene expression dataset \citep{DREAM}. The benchmark consists of five different datasets of E.Coli and yeast gene networks, each consisting of $N = 100$ variables. For every dataset, 46 time series are available, but every time series consists of only $T = 21$ time steps. We compare NAVAR to other neural approaches to Granger causality, namely componentwise-MLP (cMLP) and componentwise-LSTM (cLSTM) \citep{neuralgranger}, Temporal Causal Discovery Framework (TCDF) \citep{TCDF}, and (economy) Statistical Recurrent Units ((e)SRU) \citep{statisticalRNN} (see Related Work).

Similar to the models in \citep{statisticalRNN}, we assume a maximum lag of 2 for the MLP models and use 10 hidden units per layer. We calculate the AUROC by increasing a threshold over the causal score, where self-links are ignored in the calculation. The hyperparameters are tuned using a 80/20\% training/validation split, where we train on the first 80\% of timesteps, and select the hyperparameters with lowest mean squared error on the final 20\% time steps. The selected hyperparameters are reported in Appendix A. The hyperparameters of the other neural models are tuned in tantamount manner and can be found in \citep[Appendix~G]{statisticalRNN}. We report the average AUROC over 100 different runs of the NAVAR model. 

The results in Table \ref{table:dream} show that using deep learning to extract causal structure in time series is a non-trivial task. Our method, however, obtains the best result on all datasets. Since both the MLP and LSTM backbone outperform the other methods, we believe this is due to the imposed structure of our architecture, where the direct contributions of a variable form a more reliable indicator for causality than the methods that rely entirely on induced sparseness in the weight matrices, such as in cMLP, cLSTM, and (e)SRU. Furthermore, using permutation importance with neural networks, as in the TCDF model, is known to generate misleading conclusions~\citep{hooker2019please}. The large difference in performance between NAVAR (MLP) and NAVAR (LSTM) on the E.Coli datasets, demonstrate the benefits of exploring different backbones for different applications.

\begin{figure}[!b]
    \centering
    \includegraphics[width=0.8\linewidth]{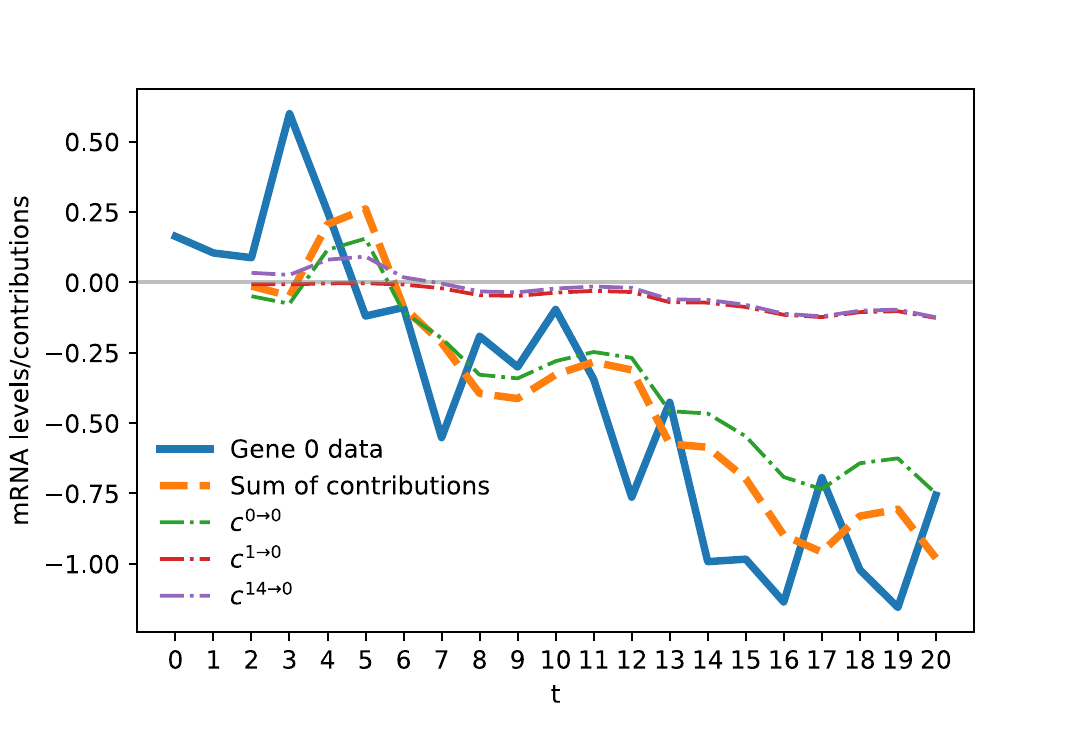}
    \caption{Example of three learned contributions to gene 0 of E.Coli 1 of the DREAM3 dataset. The original data (blue) are normalized. The final prediction is computed by summing the contributions from all genes.}
    \label{fig:geneinteractions}
\end{figure}

The linear methods are consistently outperformed by all neural methods on this dataset, which clearly indicates the importance of nonlinearity in causal structure discovery. On top of that, we also immediately obtain interpretable predictions, as shown in Figure ~\ref{fig:geneinteractions}, where we show an example of the learned causal contributions in the E.Coli 1 gene network. The model captures that the mNRA levels of gene 0 are mostly influenced by the past mNRA levels of this gene itself. However, at the end of the time series, as the levels of gene 0 go down, the influence from gene 1 and 14 pushes the gene 0 levels further down.

\section{Related work}

\subsection{Neural Methods to Causal Structure Learning}
Recently, there has been a rise in interest in applying deep learning to causal structure learning, especially within the framework of Structural Causal Models (SCM)~\citep{pearl1995causal, peters2017elements}. Research in larger graphs was limited due to a combinatorially intractable search space of possible causal graphs. A key ingredient to the solution was presented by \citet{NOTEARS}, who formulated structure learning as a continuous optimization problem. One of the key advantages of using neural networks is that one can combine the structure learning objective and the prediction objective into a single optimization problem.  
 Other methods that explore this avenue are \citep{grandag}, which extends the \citep{NOTEARS} method to nonlinear functions modeled by neural networks, while still imposing acyclicity in the causal network. Here, causal links are approximated by neural network paths. \citet{meta} and \citet{unknownint} use a meta-learning transfer objective to identify causal structures from interventional data. The structural learning objective is optimized by varying mask variables that represent the presence/missing of a causal link. \citet{kalainathan2018sam} explore the use of generative models and adversarial learning to reconstruct the causal graph.

\subsection{Causal Structure Learning for Time Series Data}
 Since there is a direct connection between differential equations and structural causal models \citep{bongers2018random}, the functioning of many complex dynamical systems can be understood in terms of causal relationships. Therefore, there has been considerable research devoted to discovering causal relationships in time series. Discovering causal relationships in these temporal settings is more straightforward than in iid data, in the sense that we can use the time-order to establish the directionality of a causal relationship. Approaches that leverage this assumption exist in many variations, such as non-parametric \citep{baek1992general, chen2006frequency}, model-based \citep{marinazzo2011nonlinear, peters2013causal},  constraint-based \citep{runge2019detecting}, and information theoretic \citep{papana2016detecting} approaches.

Despite the broad range of research in Granger causality in time series, only limited  research has applied the representational power of deep learning to this task. 
A possible reason for this is that the main challenge in causal structure learning is that the final product is the \emph{interpretation} of the dependencies between the variables, which are directly related to the causal connections. However, interpretation is known to be the Achilles heel of black-box tools such as deep learning. 

Other works that do use neural networks, such as \citep{wang2018estimating, duggento2019echo, abbasvandi2019self}, first focused on a brute force approach to estimate feature importance, where the Granger causal link $i \to j$ is estimated by the predictive power of a model for $X^{(j)}_{t}$ that includes the past of all variables, compared to a similar model where the past of the variable $X^{(i)}$ is excluded from the input. However, such an approach is not scalable when the number of variables increases. 

The Temporal Causal Discovery Framework (TCDF) \citep{TCDF} uses a attention-based (causal) convolutional neural network. They consider attention scores and introduce permutation importance to identify causal links in an additional causal validation step. Most similar to our work, \citet{neuralgranger} proposed a neural Granger causal model by using sparse component-wise MLPs (cMLP) and LSTMs (cLSTM). This approach induces sparsity on the causal links by using a hierarchical group regularization. 
\citet{statisticalRNN} use (economical) Statistical Recurrent Units to model the Granger causal dependencies, in a similar vein to the cLSTMs of \citep{neuralgranger}. Both methods use proximal gradient descent with line search to obtain interpretable results. Proximal optimization is necessary to induce \emph{exact} zeros in the weight matrices of the first layer. Exact zeros are then interpreted as a missing Granger causal link. 

In contrast, we do not limit the input features of our model, but instead, enforce interpretability directly into the architecture of our neural network by restricting the function class to produce additive features. This helps in extracting the correct causal relationships between variables, as we can directly regularize the causal summary graph instead of individual input features. Since every prediction is a sum of scalar contributions from the other variables, disentangling the effect of the different inputs becomes trivial and causal influence can be deduced intuitively.

\subsection{Neural networks as Generalised Additive Models}
In this work, we restrict the structure of the network in order to find the Granger causes of each time series. In particular, our model can be viewed as a Generalized Additive Model (GAM). In the general case, a GAM takes the form:
\begin{equation}\label{eq:gam}
    g(E[y]) = \beta + f_1(x_1) + f_2(x_2) + .. + f_n(x_n)
\end{equation}

One of the main advantages of using GAMs is that the models are considerably more interpretable than many black-box methods since the individual contributions are disentangled and evident. The benefit of assuming additive models was studied in \citep{buhlmann2014cam, peters2014causal}, but not in the context of neural networks or time series.

The use of deep learning to represent the functions $f_i$ in equation \eqref{eq:gam} was first explored in \citep{potts1999generalized} under the name Generalized Additive Neural Networks (GANNs). For a long time after, this avenue has not been explored further. Interestingly, however, in parallel to this work \citet{NAM} explored the power of Neural Additive Models (NAM) as a predictive model for tabular data with mixed data types. \citet{NAM} introduced exp-centered hidden units (ExU) to allow neural networks to easily approximate `jumpy functions', which is necessary when considering tabular data.

\section{Discussion}
We presented a neural additive extension to the autoregression framework for (Granger) causal discovery in time series, which we call NAVAR models. The choice of this architecture was guided by the success of VAR models in this context as well as by generalised additive methods as a natural extension to linear methods. We showed that neural additive models have the power to discover nonlinear relationships between time series, while they can still provide an intuitive interpretation of the learned causal interactions. Despite the fact that NAVAR does not account for higher-order interaction terms, benchmarks over a variety of datasets show that NAVAR models are more reliable than existing methods in uncovering the causal structure.

There are many interesting directions for future research. We have shown that NAVAR models already work with MLPs and LSTMs as backbone, but we can easily imagine more complex architectures, such as (dilated) CNNs and Transformers. Furthermore, it could be interesting to investigate bayesian neural networks in order to evaluate the uncertainty of a found causal model. Finally, important future work could be improvements to the model that explicitly account for unobserved confounders, non-stationarity, and contemporaneous causes.

\subsubsection{Acknowledgements} This project has received funding from the European Union’s Horizon 2020 research and innovation programme under the Marie Skłodowska-Curie grant agreement No 813114. 

\bibliography{references}

\onecolumn

\appendix
\section{Tuned Hyperparameters}
For all the tasks, we tune the batch size, learning rate, contribution penalty ($\lambda$) and weight decay ($\mu$). On the CauseMe datasets we tune the number of hidden units, whereas on the DREAM-3 dataset we tune the number of hidden layers. For computational efficiency, hyperparameters are tuned using a Tree-structured Parzen Estimator \cite{bergstra2011algorithms}. Tuned hyperparameters are provided in Tables \ref{table:causemehyp}-\ref{table:dreamhyplstm}.

\begin{table*}[!htb]
\centering
\caption{Tuned Hyperparameters of NAVAR (MLP) on the CauseMe Datasets. We Indicate the Different Variations of the ``Nonlinear VAR'' Dataset by the Number of Variables $N$ and Number of Time Steps $T$. $K$ is the Number of Lags Considered, $\lambda$ is the Contribution Penalty, and $\mu$ is the Weight Decay.}
    \centering
    \begin{tabular}{lllllllll}
        \toprule
          & \textbf{K} 

          & \textbf{\makecell{Hidden\\ Units}} & \textbf{\makecell{Layers}}  & \textbf{\makecell{Batch\\ Size}} & \textbf{\makecell{Learning\\Rate}} & $\boldsymbol{\lambda}$ & $\boldsymbol{\mu}$ \\
        \midrule
        \textbf{Tuning range} & - & [8, 128] & - & [16, 256] &  [5e-5, 5e-3] & [0, 0.5] & [1e-7, 0.5] \\
        \midrule

Nonlinear VAR \\
\quad N=3, T=300 & 5& 32 & 1 & 64 & 0.00005 & 0.1344 & 2.903e-3 \\
\quad N=5, T=300 & 5 & 16 & 1 & 64 & 0.0001 & 0.1596 & 2.420e-3 \\
\quad N=10, T=300 & 5& 128 & 1 & 64 & 0.0005 & 0.2014 & 8.557e-3 \\
\quad N=20, T=300 & 5 & 32 & 1 & 64 & 0.0002 & 0.2434 & 4.508e-3 \\
        \midrule
Climate & 2& 32 & 1 & 16 & 0.0002 & 0.3924 & 4.322e-3\\ 
Weather & 5& 32 & 1 & 64 &  0.0001 & 0.0560 & 4.903e-3 \\
River & 5 & 8 & 1 & 256 & 0.0001 & 0.1708 & 5.092e-4\\ 

        \bottomrule
    \end{tabular}

\label{table:causemehyp}

\end{table*}

\begin{table*}[!htb]
\centering
\caption{Tuned Hyperparameters of NAVAR (LSTM) on the CauseMe Datasets. We Indicate the Different Variations of the ``Nonlinear VAR'' Dataset by the Number of Variables $N$ and Number of Time Steps $T$. $K$ is the Number of Lags Considered, $\lambda$ is the Contribution Penalty, and $\mu$ is the Weight Decay.}
    \centering
    \begin{tabular}{lllllllll}
        \toprule
          & \textbf{K} 

          & \textbf{\makecell{Hidden\\ Units}} & \textbf{\makecell{Layers}}  & \textbf{\makecell{Batch\\ Size}} & \textbf{\makecell{Learning\\Rate}} & $\boldsymbol{\lambda}$ & $\boldsymbol{\mu}$ \\
        \midrule
        \textbf{Tuning range} & - & [8, 128] & - & [16, 256] &  [5e-5, 5e-3] & [0, 0.5] & [1e-7, 0.5] \\
        \midrule

Nonlinear VAR \\
\quad N=3, T=300 & 5& 16 & 1 & 64 & 0.0001 & 0.1370 & 8.952e-4 \\
\quad N=5, T=300 & 5 & 32 & 1 & 32 & 0.00005 & 0.2445 & 2.6756e-4 \\
\quad N=10, T=300 & 5& 64 & 1 & 128 & 0.0001 & 0.0784 & 7.1237e-4 \\
\quad N=20, T=300 & 5 & 128 & 1 & 64 & 0.00005 & 0.3512 & 1.901e-6 \\
        \midrule
Climate & 2& 64 & 1 & 128 & 0.0002 & 0.2334 & 6.231e-4\\ 
Weather & 5& 8 & 1 & 256 &  0.0005 & 0.0172 & 1.687e-3 \\
River & 5 & 128 & 1 & 128 & 0.001 & 0.0544 & 4.465e-4\\ 

        \bottomrule
    \end{tabular}

\label{table:causemehypLSTM}

\end{table*}

\begin{table*}[!htb]
\centering
\caption{Tuned Hyperparameters of NAVAR (MLP) on the DREAM-3 Datasets. $K$ is the Number of Lags Considered, $\lambda$ is the Contribution Penalty, and $\mu$ is the Weight Decay.}
    \centering
    \begin{tabular}{lllllllll}
        \toprule
          & \textbf{K} 
          & \textbf{\makecell{Hidden\\ Units}} & \textbf{\makecell{Layers}}  & \textbf{\makecell{Batch\\ Size}} & \textbf{\makecell{Learning\\Rate}} & $\boldsymbol{\lambda}$ & $\boldsymbol{\mu}$ \\
        \midrule
        \textbf{Tuning range} & - & - & [1, 4] & [16, 256] &  [5e-5, 5e-3] & [0, 0.5] & [1e-7, 0.5] \\
        \midrule

Ecoli1 & 2& 10 & 1 & 128 &  0.0005 & 0.1883 & 1.114e-4 \\
Ecoli2 & 2& 10 & 1 & 32 & 0.001 & 0.2011 & 1.710e-4\\ 

Yeast1 & 2 & 10 & 2  & 16 & 0.002 & 0.2697 & 1.424e-4 \\
Yeast2 & 2 & 10 & 1  & 256 & 0.0002 & 0.1563 & 2.013e-4 \\
Yeast3 & 2 & 10 & 1 & 16 & 0.0002 & 0.1559 & 1.644e-4  \\
        \bottomrule
    \end{tabular}

\label{table:dreamhyp}

\end{table*}

\begin{table}[!htb]
\centering
\caption{Tuned Hyperparameters of NAVAR (LSTM) on the DREAM-3 Datasets. $K$ is the Number of Lags Considered, $\lambda$ is the Contribution Penalty, and $\mu$ is the Weight Decay.}
    \centering
    \begin{tabular}{lllllllll}
        \toprule
          & \textbf{K} 
          & \textbf{\makecell{Hidden\\ Units}} & \textbf{\makecell{Layers}}  & \textbf{\makecell{Batch\\ Size}} & \textbf{\makecell{Learning\\Rate}} & $\boldsymbol{\lambda}$ & $\boldsymbol{\mu}$ \\
        \midrule
        \textbf{Tuning range} & - & - & - & - &  [5e-5, 5e-3] & [0, 0.5] & [1e-7, 0.5] \\
        \midrule

Ecoli1 & 21& 10 & 1 & 46 &  0.002 & 0.2208 & 1.094-5 \\
Ecoli2 & 21& 10 & 1 & 46 & 0.002 & 0.1958 & 3.233e-6\\ 

Yeast1 & 21 & 10 & 1  & 46 & 0.002 & 0.2343 & 5.309e-5 \\
Yeast2 & 21 & 10 & 1  & 46 & 0.002 & 0.2189 & 1.987-5 \\
Yeast3 & 21 & 10 & 1 & 46 & 0.002 & 0.2128 & 1.049e-5  \\
        \bottomrule
    \end{tabular}

\label{table:dreamhyplstm}

\end{table}

\clearpage

\section{ROC Curves}
\begin{figure*}[h!]
    \centering

    \includegraphics[width=0.9\linewidth]{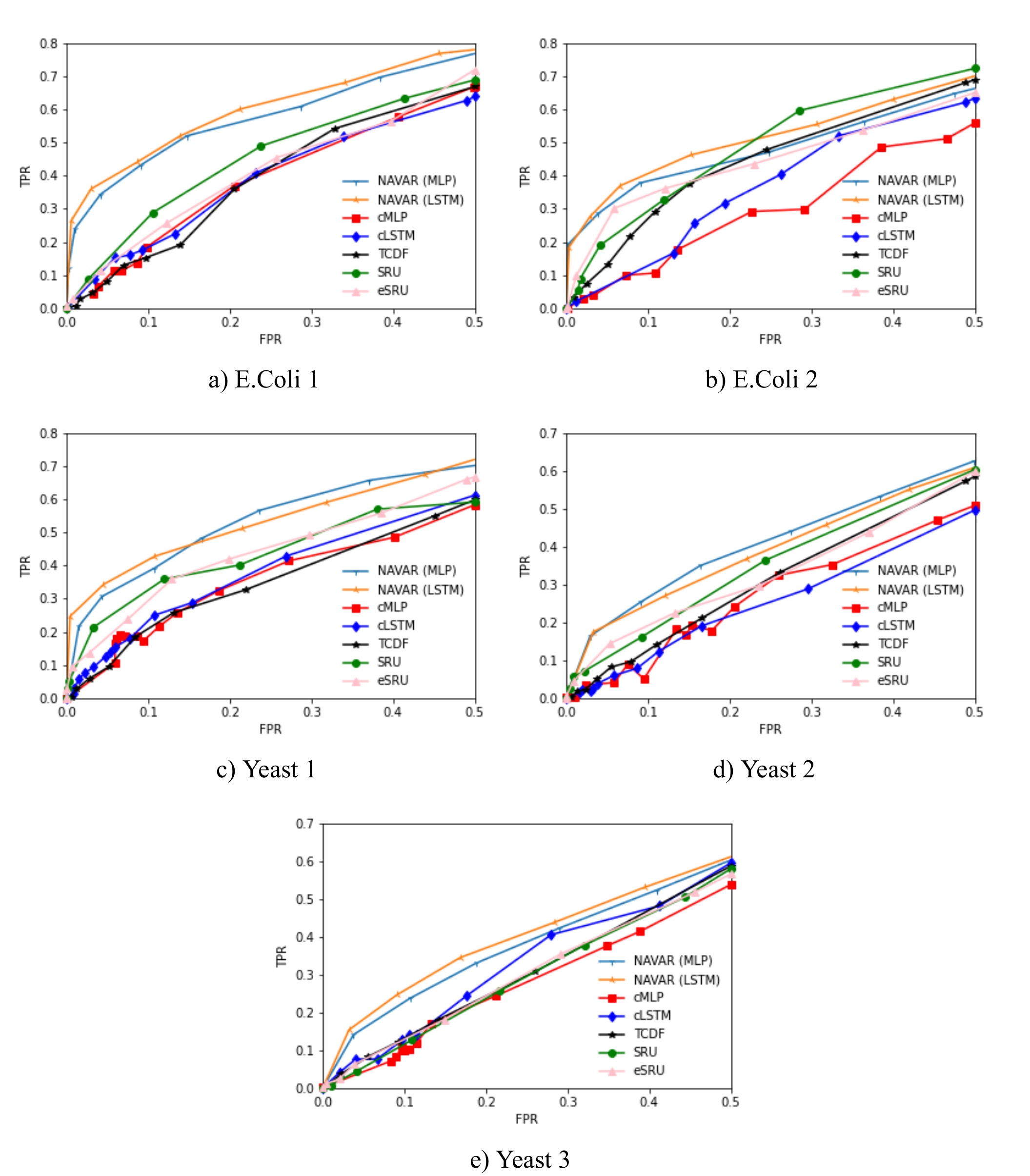}
    \caption{ROC Curves for Neural Methods on the DREAM-3 Datasets. Curves for the Methods other than NAVAR are Extracted from \cite[Figure~7]{statisticalRNN}}
    \label{fig:ROC}
\end{figure*}

The receiver operating characteristics (ROC) of the different methods are compared in Figure \ref{fig:ROC}. Here, an ROC curve represents the trade-off between the true-positive rate (TPR) and the false-positive rate (FPR) achieved by a given method while inferring the underlying pairwise causal relationships.

\end{document}